%% file: root.tex
\documentclass[letterpaper, 10 pt, conference]{ieeeconf}  

\IEEEoverridecommandlockouts                              

\usepackage{graphics} 
\usepackage{graphicx}
\usepackage{epsfig} 
\usepackage{mathptmx} 
\usepackage{times} 
\usepackage{amsmath} 
\usepackage{amssymb}  
\usepackage{hyperref}
\usepackage{adjustbox}
\usepackage[dvipsnames]{xcolor}
\usepackage{arydshln}
\usepackage{amsfonts}
\usepackage{booktabs}       
\usepackage{subcaption}
\usepackage{microtype}      
\usepackage{pifont}
\usepackage{nicefrac}       
\usepackage{multirow}
\usepackage{newtxtext, newtxmath}

\title{\LARGE \bf \input{title}}

\author{Jishnu Jaykumar P, Kamalesh Palanisamy, Yu-Wei Chao, Xinya Du, Yu Xiang
\thanks{Jishnu Jaykumar P, Kamalesh Palanisamy, Xinya Du and Yu Xiang are with the Department of Computer Science, University of Texas at Dallas, Richardson, TX 75080, USA \tt\small \{firstname.lastname\}@utdallas.edu}
\thanks{Yu-Wei Chao is with NVIDIA, Seattle, WA 98105, USA \tt\small ychao@nvidia.com}
}

\begin{document}

\maketitle
\thispagestyle{empty}
\pagestyle{empty}

\begin{abstract}
We propose a novel framework for few-shot learning by leveraging large-scale vision-language models such as CLIP~\cite{radford2021learning}. Motivated by unimodal prototypical networks for few-shot learning, we introduce \textsc{Proto-CLIP} which utilizes image prototypes and text prototypes for few-shot learning. Specifically, \textsc{Proto-CLIP} adapts the image and text encoder embeddings from CLIP in a joint fashion using few-shot examples. The embeddings from the two encoders are used to compute the respective prototypes of image classes for classification. During adaptation, we propose aligning the image and text prototypes of the corresponding classes. Such alignment is beneficial for few-shot classification due to the reinforced contributions from both types of prototypes. \textsc{Proto-CLIP} has both training-free and fine-tuned variants. We demonstrate the effectiveness of our method by conducting experiments on benchmark datasets for few-shot learning, as well as in the real world for robot perception\footnote{Project page: \url{https://irvlutd.github.io/Proto-CLIP}}.
\end{abstract}

\section{INTRODUCTION}

Building autonomous robots that can help people perform various tasks is the dream of every roboticist. Nowadays, most robots are working in factories and warehouses by performing pre-programmed repetitive tasks such as assembling and delivering. In the future, we believe that there will be intelligent robots that can perform tasks in human environments autonomously. For example, people can instruct a robot by saying ``bring me a bottle of water'' or ``wash the mug on the table'', then the robot will execute the instructions accordingly. In these scenarios, robots need to recognize objects from sensory data in order to understand the required tasks. In this work, we develop a novel few-shot learning method that can enable robots to recognize novel objects from just a few example images per object. We believe that few-shot learning~\cite{wang2020generalizing} is a promising paradigm to enable robots to recognize a large number of objects. The appeal lies in the ease of data collection---just a few example images is sufficient for teaching a robot a novel object. On the contrary, object model-based approaches build 3D models of objects and then use these 3D models~\cite{calli2015benchmarking} for object recognition. Object category-based approaches focus on recognizing category labels of objects such as 80 categories in the MSCOCO dataset~\cite{lin2014microsoft}. The limitation of model-based object recognition is the difficulty of obtaining a large number of 3D models for many objects in the real world. Current 3D scanning techniques cannot deal well with metal objects or transparent objects. For category-based object recognition, it is difficult to obtain a large number of images for each category in robotic settings. Large-scale datasets for object categories such as ImageNet~\cite{deng2009imagenet} and Visual Genome~\cite{krishna2017visual} are collected from the Internet. These Internet images are not very suitable for learning object representations for robot manipulation due to domain differences. Due to the limitations of model-based and category-based object recognition, if a robot can learn to recognize a new object from a few images of the object, it is likely to scale up the number of objects that the robot can recognize in the real world.

\begin{figure*}
\centering
\includegraphics[width=\textwidth]{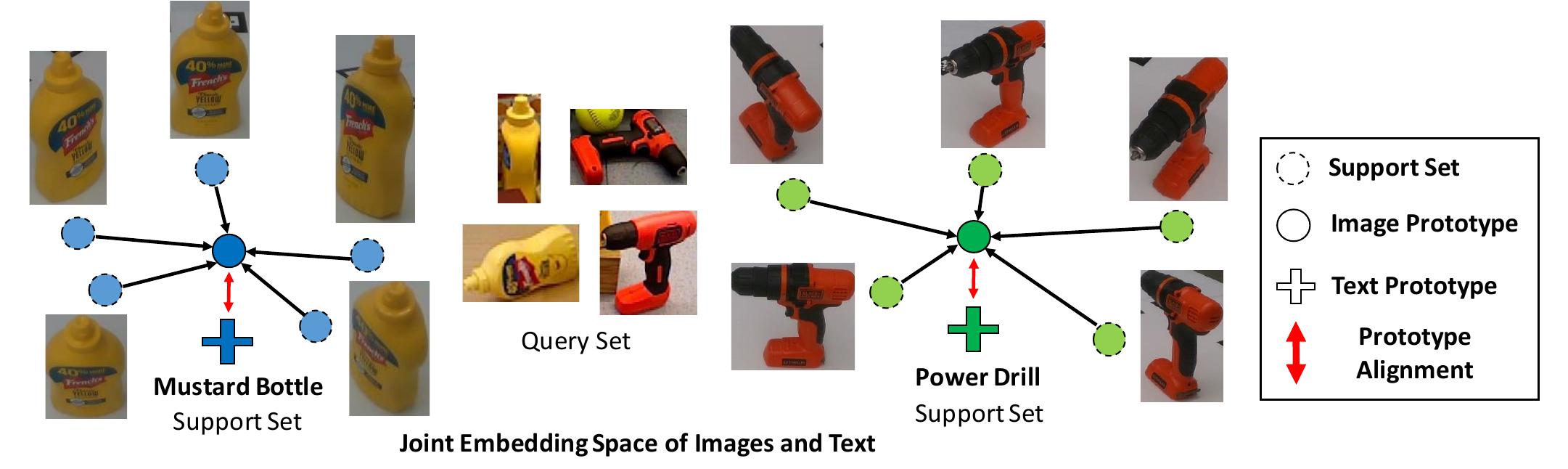}
\caption{Our \textsc{Proto-CLIP} model learns a joint embedding space of images and text, where image and text prototypes formed using support sets are learned and aligned for few-shot classification.}
\label{fig:proto_clip-intro}
\vspace{-6mm}
\end{figure*}

The main challenge in few-shot learning is how to achieve generalization with very limited training examples. Learning good visual representations is the key to achieve good performance in few-shot learning~\cite{tian2020rethinking}. Although the Internet images are quite different from robot manipulation settings, they can be used to learn powerful visual representations. Recently, the CLIP (Contrastive Language–Image Pre-training) model~\cite{radford2021learning} trained with a large number of image-text pairs from the Internet achieves promising \emph{zero-shot} image recognition performance. Using the visual and language representations from CLIP, several few-shot learning approaches~\cite{zhou2022learning,gao2021clip,tip_adapter_eccv22} are proposed to improve the zero-shot CLIP model. \cite{gao2021clip,tip_adapter_eccv22} adapt the CLIP image encoder to learn better feature representations, while \cite{zhou2022learning} learns prompts for the CLIP model. On the other hand, few-shot learning approaches are studied in the meta-learning framework~\cite{finn2017model}. These approaches are aimed at generalizing to novel classes after training. A notable method is Prototypical Network~\cite{snell2017prototypical} and its variants~\cite{triantafillou2019meta,doersch2020crosstransformers}, which demonstrate effective performance for few-shot learning. However, these methods do not leverage the powerful feature representation of CLIP.

These observations motivate us to leverage CLIP in prototypical networks for few-shot learning. We notice that existing methods for adapting CLIP models in few-shot learning adapt the image encoder~\cite{gao2021clip,tip_adapter_eccv22} or the text encoder~\cite{zhou2022learning} in CLIP. We argue that if we can use both the image encoder and the text encoder for classification and jointly adapt them using few-shot training images, we can improve the few-shot classification performance. To achieve this goal, we propose \textsc{Proto-CLIP}, a new model motivated by the traditional unimodal Prototypical Networks~\cite{snell2017prototypical}. \textsc{Proto-CLIP} utilizes image prototypes and text prototypes computed from adapted CLIP encoders for classification. In addition, we propose to align the image prototype and the text prototype of the same class during adaptation. In this way, both the image encoder and the text encoder can contribute to the classification while achieving agreement between their predictions. Fig.~\ref{fig:proto_clip-intro} illustrates the concept of learning the joint embedding space of images and text from \textsc{Proto-CLIP}.

To verify the effectiveness of \textsc{Proto-CLIP}, we have conducted experiments on commonly used benchmarks for few-shot image classification, as well as the FewSOL dataset introduced for few-shot object learning in robotic environments~\cite{p2023fewsol}. In addition, we have built a robotic system that integrates Automatic Speech Recognition (ASR), few-shot object recognition using \textsc{Proto-CLIP} and robotic grasping to demonstrate the robotic application of \textsc{Proto-CLIP}.

\section{RELATED WORK}\label{sec:related-work}

\begin{table*}[t]
\centering
\resizebox{\textwidth}{!}{
\begin{tabular}{@{}lcccc@{}}
\midrule
\textbf{Method}             & \textbf{Use Support Sets} & \textbf{Adapt Image Embedding} & \textbf{Adapt Text Embedding} & \textbf{Align Image and Text} \\ \midrule
Zero-shot CLIP~\cite{radford2021learning}           & \textcolor{red}{\ding{55}}                  & \textcolor{red}{\ding{55}}           & \textcolor{red}{\ding{55}}              & \textcolor{ForestGreen}{\ding{51}}            \\

Linear-probe CLIP~\cite{radford2021learning} & \textcolor{ForestGreen}{\ding{51}} & \textcolor{ForestGreen}{\ding{51}}           & \textcolor{red}{\ding{55}}              & \textcolor{red}{\ding{55}}  \\

CoOp~\cite{zhou2022learning} & \textcolor{ForestGreen}{\ding{51}} & \textcolor{red}{\ding{55}}           & \textcolor{ForestGreen}{\ding{51}}              & \textcolor{red}{\ding{55}}  \\

CLIP-Adapter~\cite{gao2021clip} & \textcolor{ForestGreen}{\ding{51}} &
\textcolor{ForestGreen}{\ding{51}} &
\textcolor{ForestGreen}{\ding{51}}           &  \textcolor{red}{\ding{55}} \\

Tip-Adapter~\cite{tip_adapter_eccv22} & \textcolor{ForestGreen}{\ding{51}} &
\textcolor{ForestGreen}{\ding{51}} &
\textcolor{ForestGreen}{\ding{51}}           &  \textcolor{red}{\ding{55}} \\

Sus-X~\cite{udandarao2022susx} & \textcolor{ForestGreen}{\ding{51}} &
\textcolor{ForestGreen}{\ding{51}} &
\textcolor{red}{\ding{55}}           &  \textcolor{red}{\ding{55}} \\ \hline
\rule{0pt}{10pt}
\textbf{\textsc{Proto-CLIP} (Ours)} & \textcolor{ForestGreen}{\ding{51}} &
\textcolor{ForestGreen}{\ding{51}} &
\textcolor{ForestGreen}{\ding{51}}         &  \textcolor{ForestGreen}{\ding{51}} \\

\bottomrule
\end{tabular}}
\caption{ Comparison of our proposed method with the existing CLIP-based few-shot learning methods. ``Use Support Sets'' indicates if a method uses support training sets for fine-tuning. ``Adapt Image/Text Embedding'' indicates if a method adapts the image/text embeddings obtained from CLIP. ``Align Image and Text'' indicates if a method \textit{specifically} aligns images and text in the feature space.}
\label{tab:related-work-summary}
\vspace{-6mm}
\end{table*}

In the context of image recognition, few-shot learning indicates using a few images per image category. The problem is usually formulated as ``$N$-way, $K$-shot'', i.e., $N$ classes with $K$ images per class. In the traditional image classification setup, these $NK$ images are used as training images. Once a model is trained, it can be used to test images among $N$ classes. Recent CLIP-based few-shot learning methods fall into this setting.\\

\textbf{CLIP-based Few-Shot Learning.} The CLIP~\cite{radford2021learning} model applies contrastive learning to image-text pairs from the Internet. It consists of an image encoder and a text encoder for the extraction of features from images and text, respectively. Its training objective is to maximize the similarity between the corresponding image and text in a pair in a high-dimensional joint feature space. After training, CLIP can be used for zero-shot image classification by comparing image features with text embeddings of novel class names. This model is denoted as zero-shot CLIP. When a few training images are available for each class, several approaches are proposed to improve zero-shot CLIP. The linear-probe CLIP model~\cite{radford2021learning} trains a logistic regression classifier using CLIP image features. CoOp~\cite{zhou2022learning} proposes to use learnable vectors as a prompt for the CLIP text encoder for few-shot learning. CLIP-Adapter~\cite{gao2021clip} learns two layers of linear transformations on top of the image encoder and the text encoder with residual connections, respectively, to adapt CLIP features for few-shot learning. Tip-Adapter~\cite{tip_adapter_eccv22} builds a key-value cache model, where keys are CLIP image features and values are one-hot vectors of the class labels. Given a query image, its image feature is compared with the cache keys to combine the value labels for classification. Tip-Adapter can also fine-tune the keys by treating them as learnable parameters, which further improves the few-shot classification accuracy. Sus-X~\cite{udandarao2022susx} leverages the power of Stable Diffusion~\cite{rombach2022highStableDiffusion} to create support sets and aims to address the issue of uncalibrated intra-modal embedding distances in TIP-Adapter~\cite{tip_adapter_eccv22} by utilizing inter-modal distances as a connecting mechanism. Table~\ref{tab:related-work-summary} compares our proposed method with existing CLIP-model-based few-shot learning methods. By using the image prototypes and text prototypes for classification, our method can adapt both the image embeddings and text embeddings from CLIP. In addition, the model aligns the image prototypes and the text prototypes, which serves as a regularization term in adapting the feature embeddings. We empirically verify our model by conducting experiments on benchmark datasets for few-shot learning.

\textbf{Meta-learning-based Few-Shot Learning.} In parallel with these efforts to adapt CLIP for few-shot learning, meta-learning-based approaches are also proposed for few-shot learning. While previous CLIP-based models are tested on the same classes in training, the focus here is to learn a model on a set of training classes $\mathcal{C}_{train}$ that can generalize to novel classes $\mathcal{C}_{test}$ in testing. Each class contains a support set and a query set. During training, the class labels for both sets are available. During testing, only the class labels of the support set are available, and the goal is to estimate the labels of the query set. Meta-learning-based approaches train a meta-learner with the training classes $\mathcal{C}_{train}$ that can be adapted to the novel classes $\mathcal{C}_{test}$ using their support sets. Non-episodic approaches use all the data in $\mathcal{C}_{train}$ for training such as $k$-NN and its `Finetuned' variants~\cite{gidaris2018dynamic,qi2018low,chen2019closer,tian2020rethinking}. Episodic approaches construct episodes, i.e., a subset of the training classes, to train the meta-learner. Representative episodic approaches include Prototypical Networks~\cite{snell2017prototypical}, Matching Networks~\cite{vinyals2016matching}, Relation Networks~\cite{sung2018learning}, Model Agnostic Meta-Learning (MAML)~\cite{finn2017model}, Proto-MAML~\cite{triantafillou2019meta} and CrossTransformers~\cite{doersch2020crosstransformers}. The \textsc{Meta-Dataset}~\cite{triantafillou2019meta} was introduced to benchmark few-shot learning methods in this setting. In this work, we consider training and testing in the same classes following previous CLIP-based few-shot learning methods~\cite{zhou2022learning,gao2021clip,tip_adapter_eccv22}.

\section{METHOD}\label{sec: method}
We consider the $N$-way $K$-shot classification problem. In few-shot settings, $K \ll N$. The image set with class labels is considered as the \emph{support set}: $\mathcal{S} = \{ \mathbf{x}_i^s, y_i^s \}_{i=1}^M$, where $\mathbf{x}_i^s$ denotes a support image, $y_i^s \in \{1, 2, \ldots, N \}$ denotes the class label of the support image, and $M$ is the size of the support set. In $N$-way $K$-shot settings, $M = NK$. The goal of few-shot classification is to classify the \emph{query set} $\mathcal{Q} = \{ \mathbf{x}_j^q \}_{j=1}^{L}$, i.e., $L$ test images without class labels. Specifically, we want to estimate the conditional probability $P(y = k | \mathbf{x}^q, \mathcal{S})$ that models the probability distribution of the class label $y$ given a query image $\mathbf{x}^q$ and the support set $\mathcal{S}$.


\begin{figure*}[t]
\centering
\includegraphics[width=\textwidth]{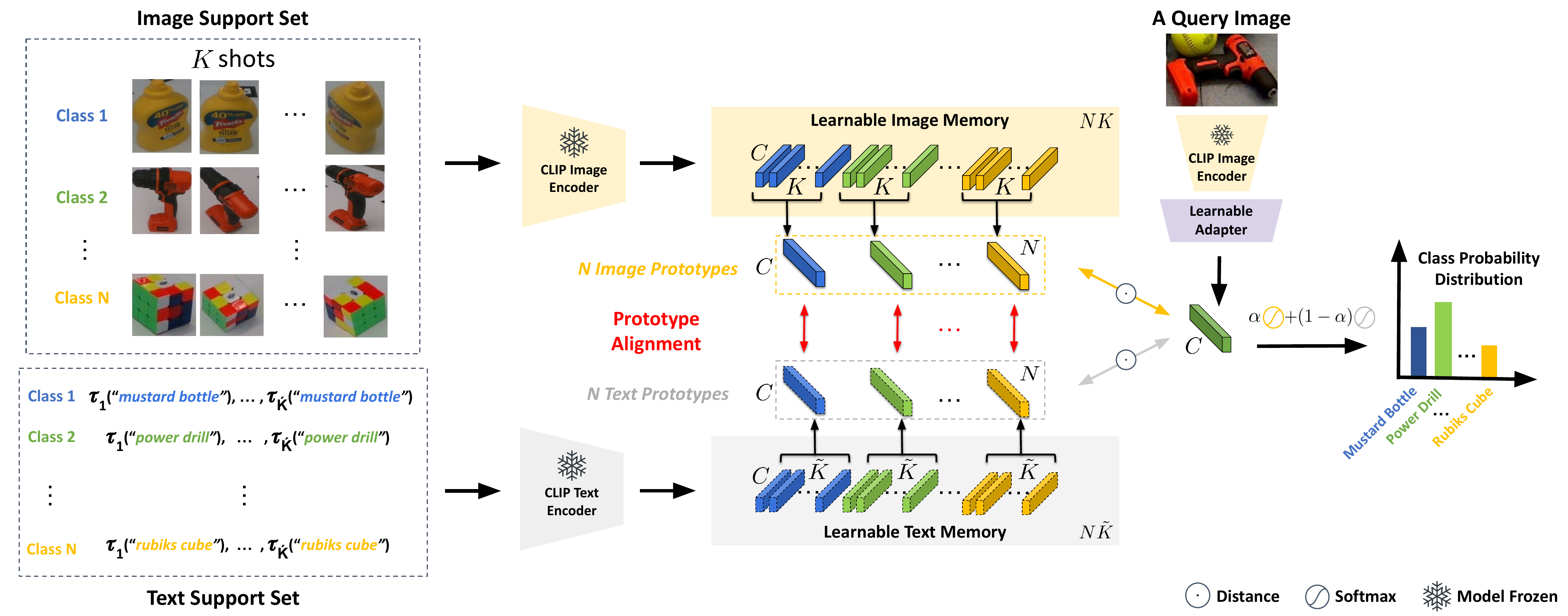}
\caption{Overview of our proposed \textsc{Proto-CLIP} model. The image memory, the text memory and the adapter network are learned. Given a class name, $\tau_i$ returns the $i^{th}$ out of $\tilde{K}$ predefined text prompts.}
\label{fig:proto_clip}
\vspace{-3mm}
\end{figure*}

\textbf{Our \textsc{Proto-CLIP} model (Fig.~\ref{fig:proto_clip})}. We propose to leverage both the image encoder and the text encoder in the CLIP model~\cite{radford2021learning} to estimate the conditional probability of class label as
\begin{equation} \label{eq:prob}
    P(y = k | \mathbf{x}^q, \mathcal{S}) = \alpha \underbrace{ P(y = k | \mathbf{x}^q, \mathcal{S}_x) }_{\text{image probability}} + (1 - \alpha) \underbrace{ P(y = k | \mathbf{x}^q, \mathcal{S}_y) }_{\text{text probability}},
\end{equation}
where $\mathcal{S}_x = \{ \mathbf{x}_i^s\}_{i=1}^M$ and $\mathcal{S}_y = \{ y_i^s\}_{i=1}^M$ denote the image set and the label set of the support set $\mathcal{S}$, respectively, and $\alpha \in [0, 1]$ is a hyper-parameter to combine the two probabilities. To model the probability distributions conditioned on $\mathcal{S}_x$ or $\mathcal{S}_y$, we leverage the prototypical networks~\cite{snell2017prototypical}:
\begin{eqnarray}
 \label{eq:proto1}
    P(y = k | \mathbf{x}^q, \mathcal{S}_x) = \frac{\exp(- \beta \| g_{\mathbf{w}_1} (\mathbf{x}^q) - \mathbf{c}_k^x \|_2^2)}{\sum_{k'=1}^N \exp(-\beta \| g_{\mathbf{w}_1}(\mathbf{x}^q) - \mathbf{c}_{k'}^x \|_2^2 )}, \\ \label{eq:proto2}
    P(y = k | \mathbf{x}^q, \mathcal{S}_y) = \frac{\exp(-\beta \| g_{\mathbf{w}_1}(\mathbf{x}^q) - \mathbf{c}_k^y \|_2^2)}{\sum_{k'=1}^N \exp(-\beta \| g_{\mathbf{w}_1}(\mathbf{x}^q) - \mathbf{c}_{k'}^y \|_2^2 )},
\end{eqnarray}
where $g_{\mathbf{w}_1}(\cdot)$ denotes the CLIP image encoder plus an adapter network with learnable parameters $\mathbf{w}_1$ used to compute the feature embeddings of query images. The CLIP image encoder is pretrained and then frozen. $\mathbf{c}_k^x$ and $\mathbf{c}_k^y$ are the ``prototypes'' for class $k$ computed using images and text, respectively. $\beta \in \mathbb{R}^{+}$ is a hyperparameter to sharpen the probability distributions. We have the prototypes as 
\begin{equation}
\mathbf{c}_k^x = \frac{1}{M_k} \sum_{y_i^s = k} \phi_\text{Image} (\mathbf{x}_i^s)
\end{equation}
\begin{equation}
 \; \; \mathbf{c}_k^y = \frac{1}{\tilde{M_k}} \sum_{j = 1}^{\tilde{M_k}} \phi_\text{Text}( \text{Prompt}_j (y_i^s = k)),
\end{equation} 


where $M_k$ is the number of examples with label $k$, and $\tilde{M_k}$ is the number of prompts for label $k$. To compute text embeddings, we can either directly input the class names such as ``mug'' and ``plate'' into the text encoder, or convert the class names to phrases such as ``a photo of mug'' and ``a photo of plate'' and then input the phrases into the text encoder. These phrases are known as \emph{prompts} of the vision-language models. We can use multiple prompts for each class label. $\phi_\text{Image}(\mathbf{x}_i^s)$ and $\phi_\text{Text}( \text{Prompt}_j (y_i^s = k))$ denote the image embedding and the $j$th text embedding of the image-label pair $(\mathbf{x}_i^s, y_i^s)$ computed using the CLIP image encoder and the text encoder, respectively. These embeddings with dimension $C$ of the support set form the image memory and the text memory, as shown in Fig.~\ref{fig:proto_clip}. They are learnable embedding vectors initialized by the computed embeddings using the CLIP image encoder and text encoder. We use $\mathbf{c}_k^x$ and $\mathbf{c}_k^y$ to denote the mean of the embeddings of the images and the prompts for class $k$, respectively. Since the image embeddings and the text embeddings are of the same dimension, we can compute the distance between the text prototype $\mathbf{c}_k^y$ and the image embedding $g_{\mathbf{w}_1}(\mathbf{x}^q)$ in Eq.~\ref{eq:proto2}. As we can see, our model leverages prototypical networks with image encoder and text encoder from CLIP. We name it ``\textsc{Proto-CLIP}''.

\begin{figure*}[h]
\centering
\includegraphics[width=\textwidth]{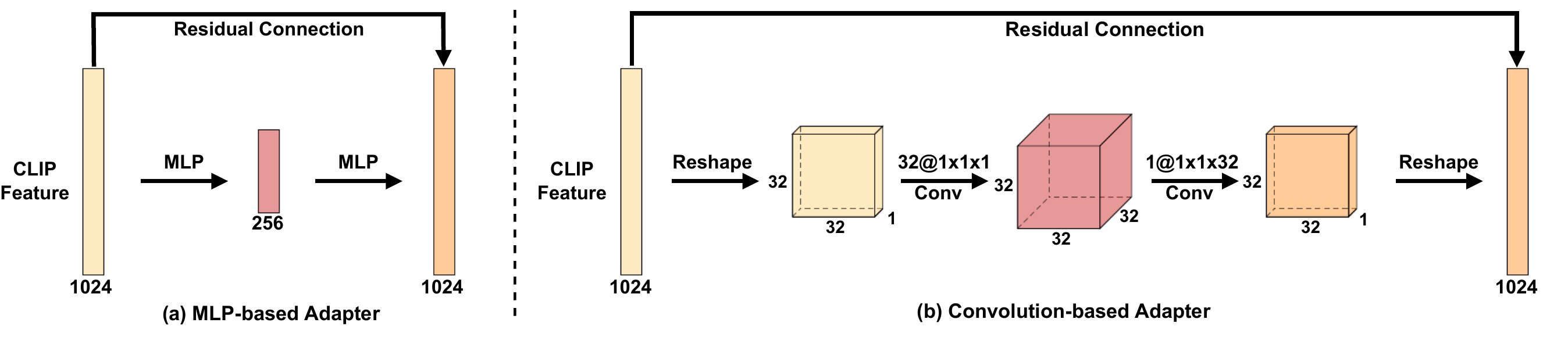}
\caption{Two designs of the adapters. (a) A Multi-layer perceptron-based adapter as in~\cite{gao2021clip}. (b) A convolution-based adapter that we introduce. The feature dimension is for CLIP ResNet50 backbone.}
\label{fig:adapter}
\vspace{-8mm}
\end{figure*}

\textbf{Learning the memories and the adapter.} During training, we can construct a support set $\mathcal{S} = \{ \mathbf{x}_i^s, y_i^s \}_{i=1}^M$ and a query set with ground truth labels $\mathcal{Q} = \{ \mathbf{x}_j^q, y_j^q \}_{j=1}^L$. Then we can use $\mathcal{S}$ and $\mathcal{Q}$ to learn the weights in \textsc{Proto-CLIP}. First, the support set is used to initialize the image memory $\mathbf{W}_{\text{image}}$ and the text memory $\mathbf{W}_{\text{text}}$. Second,  the weights in the adapter network applied to the query images $g_{\mathbf{w}_1}(\cdot)$ need to be learned. Fig.~\ref{fig:adapter} shows two designs of the adapter network, i.e., an MLP-based adapter as in~\cite{gao2021clip} and a convolution-based adapter that we introduce. The convolution-based adapter has fewer weights to learn compared to the MLP-based one. We found that the two adapters have their own advantages on different datasets in our experiments. Finally, motivated by the CLIP-Adapter~\cite{gao2021clip}, we do not fine-tune the weights in the image encoder and text encoder by freezing these weights during training. In this way, we can reuse the weights of CLIP trained on a large number of image-text pairs and adapt the image embeddings and the text embeddings.

\textbf{Loss Functions.} The first loss function is the negative log-probability of the true label for a query image: $\mathcal{L}_1(\mathbf{W}_{\text{image}}, \mathbf{W}_{\text{text}}, \mathbf{w}_1) = - \log  P(y^q = k | \mathbf{x}^q, \mathcal{S})$, where $P(y^q = k | \mathbf{x}^q, \mathcal{S})$ is defined in Eq.~\ref{eq:prob}. Minimizing $\mathcal{L}_1$ learns the weights to classify the query images correctly. Second, we propose aligning the image prototypes and the text prototypes in training. Let $\{ \mathbf{c}_1^x, \mathbf{c}_2^x, \ldots, \mathbf{c}_N^x \}$ be the image prototypes computed from the image embeddings for the $N$ classes and $\{ \mathbf{c}_1^y, \mathbf{c}_2^y, \ldots, \mathbf{c}_N^y \}$ be the corresponding text prototypes. We would like to learn the model weights such that $\mathbf{c}_k^x$ is close to $\mathbf{c}_k^y$ and far from other prototypes in the embedding space. We utilize the InfoNCE loss for contrastive learning~\cite{oord2018representation}:
\begin{equation}\label{eq:align-l2}
    \mathcal{L}_2^k(\mathbf{c}_k^x, \{ \mathbf{c}_{k'}^y \}_{k'=1}^N) = - \log \frac{\exp(\mathbf{c}_k^x \cdot \mathbf{c}_k^y)}{\sum_{k'=1}^N \exp(\mathbf{c}_k^x \cdot \mathbf{c}_{k'}^y) }
\end{equation}
\begin{equation}\label{eq:align-l3}
    \mathcal{L}_3^k(\mathbf{c}_k^y, \{ \mathbf{c}_{k'}^x \}_{k'=1}^N) = - \log \frac{\exp(\mathbf{c}_k^y \cdot \mathbf{c}_k^x)}{\sum_{k'=1}^N \exp(\mathbf{c}_k^y \cdot \mathbf{c}_{k'}^x)}
\end{equation}
for $k=1,\dots,N$, where $\cdot$ indicates dot-product. Here, $\mathcal{L}_2^k(\mathbf{c}_k^x, \{ \mathbf{c}_{k'}^y \}_{k'=1}^N)$ compares an image prototype $\mathbf{c}_k^x$ with the text prototypes $\{ \mathbf{c}_{k'}^y \}_{k'=1}^N$, while $\mathcal{L}_3^k(\mathbf{c}_k^y, \{ \mathbf{c}_{k'}^x \}_{k'=1}^N)$ compares a text prototype $\mathbf{c}_k^y$ with the image prototypes $\{ \mathbf{c}_{k'}^x \}_{k'=1}^N$. In this way, we can align the image prototypes and the text prototypes for the $N$ classes. This alignment can facilitate classification, since the class conditional probabilities are computed using the image prototypes and the text prototypes as in Eqs.~\ref{eq:proto1} and \ref{eq:proto2}. The total loss function for training is:
\begin{align} \label{eq: total-loss}
\begin{split}
    \mathcal{L} &= - \frac{1}{L} \sum_{j=1}^L \log  P(y_j^q = k | \mathbf{x}_j^q, \mathcal{S}) \\
    &  + \frac{1}{N} \sum_{k=1}^N \big( \mathcal{L}_2^k(\mathbf{c}_k^x, \{ \mathbf{c}_{k'}^y \}_{k'=1}^N) + \mathcal{L}_3^k(\mathbf{c}_k^y, \{ \mathbf{c}_{k'}^x \}_{k'=1}^N) \big)
    \end{split}
\end{align}

    

for a query set $\mathcal{Q} = \{ \mathbf{x}_j^q, y_j^q \}_{j=1}^L$.  Following previous CLIP-based few-shot learning methods~\cite{zhou2022learning,gao2021clip,tip_adapter_eccv22}, the support set and the query set are the same during training in our experiments, i.e., $\mathcal{S} = \mathcal{Q}$ meaning any of the support samples can act as a query sample during training.

\begin{figure*}
    \centering
    \includegraphics[width=\linewidth]{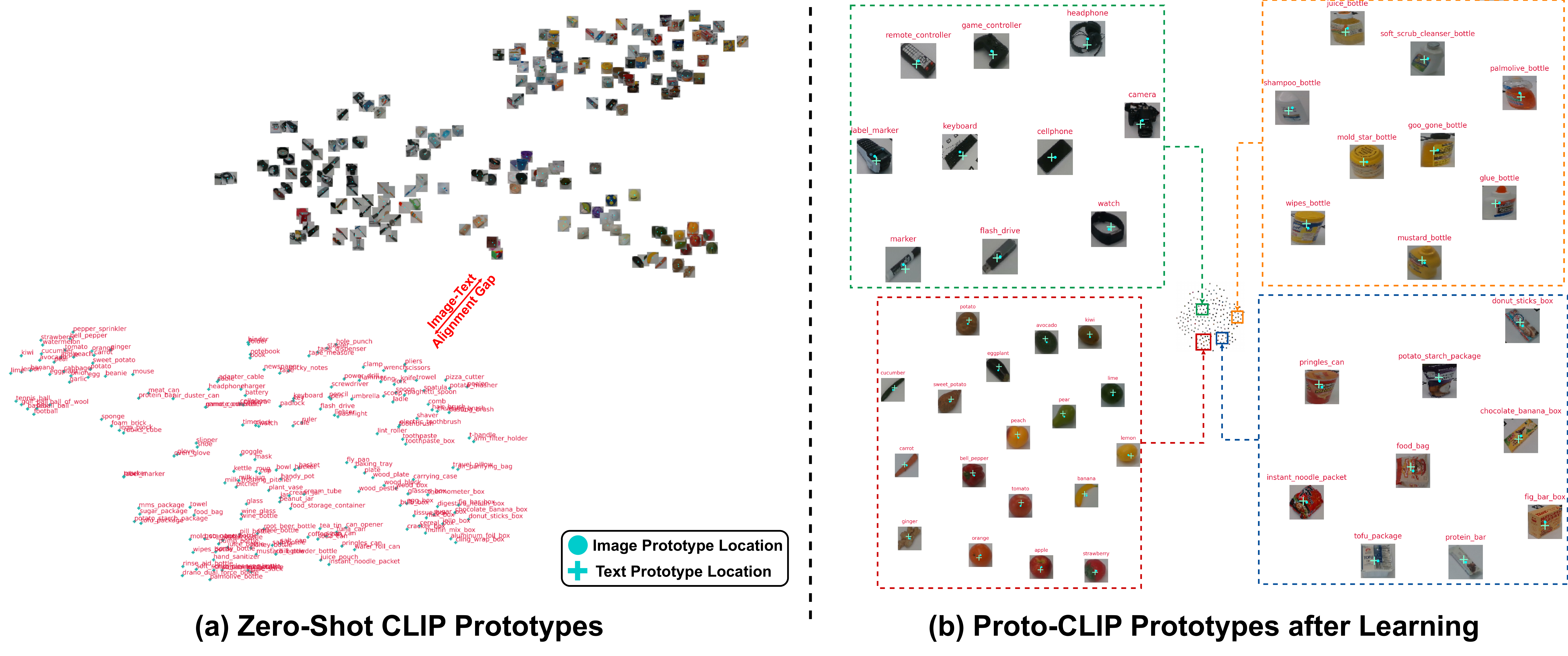}
    \vspace{-2mm}
    \caption{ Barnes-Hut t-SNE visualization~\cite{van2014accelerating} using the FewSOL dataset~\cite{p2023fewsol}. (a) Image and text prototypes from zero-shot CLIP, which are not aligned. (b) Aligned image and text prototypes from \textsc{Proto-CLIP-$F$}.}
    \vspace{-7mm}
    \label{fig:tsne}
\end{figure*}

\section{EXPERIMENTS}\label{sec:experiments}

\textbf{Datasets and Evaluation Metric.}
Following previous CLIP-based few-shot learning methods~\cite{zhou2022learning,gao2021clip,tip_adapter_eccv22}, we conduct experiments on the following datasets for evaluation: ImageNet~\cite{deng2009imagenet}, StandfordCars~\cite{krause20133d}, UCF101~\cite{soomro2012ucf101},
Caltech101~\cite{fei2004learning}, Flowers102~\cite{nilsback2008automated}, SUN397~\cite{xiao2010sun}, DTD~\cite{cimpoi2014describing}, EuroSAT~\cite{helber2019eurosat}, FGVCAircraft~\cite{maji2013fine}, OxfordPets~\cite{parkhi2012cats}, and Food101~\cite{bossard2014food}. In addition, we also include the FewSOL dataset~\cite{p2023fewsol} recently introduced for few-shot object recognition in robotic environments in order to improve object classification for robot manipulation tasks. In the $N$-way $K$-shot classification setting, $K$ images for each class will be sampled from each dataset for training. A validation set of each dataset is reserved for hyper-parameter tuning, and a test set is used for evaluation. We evaluate using test set classification accuracy, as in related works.

\textbf{Choosing the Hyper-parameters: $\alpha$ and $\beta$.} From the experiments, we found that the two hyper-parameters $\alpha$ in Eq.~\ref{eq:prob} and $\beta$ in Eq.~\ref{eq:proto1} and Eq.~\ref{eq:proto2} play a critical role in classification accuracy. Therefore, for each dataset, we conducted a grid search of the two parameters using the validation set. Then we finalize their values for all the runs in our experiments.

\textbf{\textsc{Proto-CLIP} Variants.} i) ``\textsc{Proto-CLIP}'': we do not train the image memory and the text memory and do not use any adapter in \textsc{Proto-CLIP} (Fig.~\ref{fig:proto_clip}), we directly run inference using the pre-trained CLIP features. We term this variant the ``training-free" version because it does not require training. This offers a convenient way to quickly test new datasets without the complexities of training, although it comes with the caveat of potential misalignment between visual and textual features. ii) ``\textsc{Proto-CLIP}-$F$'': we train the image memory and/or the text memory with the adapter. During training, for all the query images, we precompute their CLIP image features and directly use these stored features for training. This variant can be trained more quickly w.r.t. the following variant. Therefore, we use it for our ablation studies. iii) ``\textsc{Proto-CLIP}-$F$-$Q^T$'': During training, for each query image, we apply random data augmentation operations such as cropping and horizontal flip. Then we compute CLIP image features for the transformed query images during training.

\vspace{-0.5mm}
\subsection{Ablation Studies} \label{sec: ablation-study}
\vspace{-0.5mm}
\textbf{Adapter Types and Learnable Text Memory.} Since the 12 datasets have different characteristics, we found that varying adapter types and whether to learn the text memory or not affect performance. Table~\ref{table:abl-study-adapter-k-16} summarizes the result of this ablation study. Visual data plays a crucial role in image recognition when compared to textual information. Therefore, visual memory keys are consistently trained, regardless of the circumstances. The architectures of the MLP-based adapter and the convolution-based adapter are illustrated in Fig.~\ref{fig:adapter}. ``2xConv'' indicates using 2 convolution layers as shown in Fig.~\ref{fig:adapter}, while ``3xConv'' uses 3 convolution layers in the adapter where we add a $32 @ 3\times3\times 32$ convolution layer in the middle.  By checking the best accuracy for each dataset, we can observe that there is no consensus on which adapter and trainable text memory setup to use among these datasets. Therefore, we select the best configuration on the adapter and learnable text memory for each dataset in the following experiments. Learning both image memory and text memory can help to yield aligned image-text prototypes. Fig.~\ref{fig:tsne} visualizes the image-text prototypes in the FewSOL dataset~\cite{p2023fewsol} before and after training. For \textsc{Proto-CLIP-$F$}, unless specified otherwise, both the adapter and the visual memory keys are trained in all scenarios.

\begin{table*}[h]
\centering
\resizebox{\textwidth}{!}{%
\begin{tabular}{l|cccccccccccc}
\hline
\textbf{Dataset} & \textbf{ImageNet} & \textbf{FGVC} & \textbf{Pets} & \textbf{Cars} & \textbf{EuroSAT} & \textbf{Caltech101} & \textbf{SUN397} &  \textbf{DTD} & \textbf{Flowers} & \textbf{Food101} & \textbf{UCF101} & \textsc{\textbf{FewSOL}}\\
\textbf{\# classes} & \textbf{1000}  & \textbf{100} & \textbf{37} &  \textbf{196} & \textbf{10} & \textbf{100} & \textbf{397} & \textbf{47}  & \textbf{102} & \textbf{101} & \textbf{101} & \textbf{52} \\ \hline
Zero-shot CLIP~\cite{radford2021learning} & 60.33 & 17.10 & 85.83 & 55.74 & 37.52 & 85.92 & 58.52 & 42.20 & 66.02 & 77.32 & 61.35 & 25.91 \\ \hline
\multicolumn{13}{c}{\textbf{1 shot}} \\ \hline
Linear-Probe CLIP~\cite{radford2021learning} & 22.07 & 12.89 & 30.14 & 24.64 & 51.00 & 70.62 & 32.80 & 29.59 & 58.07 & 30.13 & 41.43 &  -  \\
CLIP-Adapter~\cite{gao2021clip} & 61.20 & 17.49 & 85.99 & 55.13 & 61.40 & 88.60 & 61.30 & 45.80 & 73.49 & 76.82 & 62.20 &  -  \\
CoOp~\cite{zhou2022learning} & 57.15 & 9.64 & 85.89 & 55.59 & 50.63 & 87.53 & 60.29 & 44.39 & 68.12 & 74.32 & 61.92 &  -  \\
Tip~\cite{tip_adapter_eccv22} & 60.70 & 19.05 & 86.10 & 57.54 & 54.38 & 87.18 & 61.30 & 46.22 & 73.12 & 77.42 & 62.60 &  27.30  \\
Tip-F~\cite{tip_adapter_eccv22} & \textbf{61.13} & \textbf{20.22} & \textbf{87.00} & \textbf{58.86} & 59.53 & \textbf{89.33} & \textbf{62.50} & \textbf{49.65} & \textbf{79.98} & \textbf{77.51} & \textbf{64.87} &  \textbf{27.91}  \\

\textsc{Proto-CLIP} & 60.31 & 19.59 & 86.10 & 57.29 & 55.53 & 87.99 & 60.81 & 46.04 & 76.98 & 77.36 & 63.15 & 27.09 \\
\textsc{Proto-CLIP}-$F$ & 60.32& 19.50& 85.72& 57.34& 54.93& 88.07& 60.83& 35.64& 77.47& 77.34& 63.07& 22.22 \\
 
\textsc{Proto-CLIP}-$F$-${Q^{T}}$ & 59.12& 16.26& 83.62& 52.77& \textbf{61.95}& 88.48& 61.43& 32.27& 68.53& 75.16& 62.44& 21.65\\\hline

\multicolumn{13}{c}{\textbf{2 shots}} \\ \hline
Linear-Probe CLIP~\cite{radford2021learning} & 31.95 & 17.85 & 43.47 & 36.53 & 61.58 & 78.72 & 44.44 & 39.48 & 73.35 & 42.79 & 53.55 &  -  \\
CLIP-Adapter~\cite{gao2021clip} & 61.52 & 20.10 & 86.73 & 58.74 & 63.90 & 89.37 & 63.29 & 51.48 & 81.61 & 77.22 & 67.12 &  -  \\
CoOp~\cite{zhou2022learning} & 57.81 & 18.68 & 82.64 & 58.28 & 61.50 & 87.93 & 59.48 & 45.15 & 77.51 & 72.49 & 64.09 &  -  \\
Tip~\cite{tip_adapter_eccv22} & 60.96 & 21.21 & 87.03 & 57.93 & 61.68 & 88.44 & 62.70 & 49.47 & 79.13 & 77.52 & 64.74 &  26.22  \\
Tip-F~\cite{tip_adapter_eccv22} & \textbf{61.69} & \textbf{23.19} & 87.03 & \textbf{61.50} & \textbf{66.15} & \textbf{89.74} & 63.64 & \textbf{53.72} & 82.30 & \textbf{77.81} & 66.43 &  27.43  \\


\textsc{Proto-CLIP} &  60.64& 22.14& \textbf{87.38}& 60.01& 64.89& 89.05& 63.12& 51.06& 83.39& 77.34& 67.46& \textbf{28.35} \\
\textsc{Proto-CLIP}-$F$ & 60.64& 22.14& \textbf{87.38}& 60.04& 64.86& 89.09& 63.20& 49.88& \textbf{83.52}& 77.34& 67.49& 26.17 \\
\textsc{Proto-CLIP}-$F$-${Q^{T}}$ & 60.48& 20.01& 85.28& 60.02& 63.59& 89.49& \textbf{65.46}& 45.69& 81.20& 76.15& \textbf{68.83}& 25.91 \\\hline


\multicolumn{13}{c}{\textbf{4 shots}} \\ \hline
Linear-Probe CLIP~\cite{radford2021learning} & 41.29 & 23.57 & 56.35 & 48.42 & 68.27 & 84.34 & 54.59 & 50.06 & 84.80 & 55.15 & 62.23 &  -  \\
CLIP-Adapter~\cite{gao2021clip} & 61.84 & 22.59 & 87.46 & 62.45 & 73.38 & 89.98 & 65.96 & 56.86 & 87.17 & 77.92 & 69.05 &  -  \\
CoOp~\cite{zhou2022learning} & 59.99 & 21.87 & 86.70 & 62.62 & 70.18 & 89.55 & 63.47 & 53.49 & 86.20 & 73.33 & 67.03 &  -  \\
Tip~\cite{tip_adapter_eccv22} & 60.98 & 22.41 & 86.45 & 61.45 & 65.32 & 89.39 & 64.15 & 53.96 & 83.80 & 77.54 & 66.46 &  28.70  \\
Tip-F~\cite{tip_adapter_eccv22} & \textbf{62.52} & 25.80 & \textbf{87.54} & 64.57 & 74.12 & 90.56 & 66.21 & \textbf{57.39} & 88.83 & \textbf{78.24} & \textbf{70.55} &  29.13  \\


\textsc{Proto-CLIP} & 61.30& 23.25& 87.19& 63.33& 68.67& 89.57& 65.51& 55.91& 88.23& 77.58& 69.50& 29.13 \\
\textsc{Proto-CLIP}-$F$ & 61.30& 23.31& 86.95& 63.34& 68.52& 89.62& 65.57& 57.21& 88.27& 77.58& 69.55& 27.09 \\
\textsc{Proto-CLIP}-$F$-${Q^{T}}$ & 61.80& \textbf{27.63}& 87.11& \textbf{66.24}& \textbf{80.64}& \textbf{91.81}& \textbf{68.09}& 56.86& \textbf{89.85}& 76.94& 70.16& \textbf{30.30} \\\hline

\multicolumn{13}{c}{\textbf{8 shots}} \\ \hline
Linear-Probe CLIP~\cite{radford2021learning} & 49.55 & 29.55 & 65.94 & 60.82 & 76.93 & 87.78 & 62.17 & 56.56 & 92.00 & 63.82 & 69.64 &  -  \\
CLIP-Adapter~\cite{gao2021clip} & 62.68 & 26.25 & 87.65 & 67.89 & 77.93 & 91.40 & 67.50 & 61.00 & 91.72 & 78.04 & 73.30 &  -  \\
CoOp~\cite{zhou2022learning} & 61.56 & 26.13 & 85.32 & 68.43 & 76.73 & 90.21 & 65.52 & 59.97 & 91.18 & 71.82 & 71.94 &  -  \\
Tip~\cite{tip_adapter_eccv22} & 61.45 & 25.59 & 87.03 & 62.93 & 67.95 & 89.83 & 65.62 & 58.63 & 87.98 & 77.76 & 68.68 &  29.22  \\
Tip-F~\cite{tip_adapter_eccv22} & 64.00 & 30.21 & 88.09 & 69.25 & 77.93 & 91.44 & 68.87 & 62.71 & 91.51 & \textbf{78.64} & 74.25 &  32.43  \\

\textsc{Proto-CLIP} &  62.12& 27.63& 88.04& 64.93& 69.42& 90.22& 67.37& 59.34& 92.08& 77.90& 71.08& 29.83 \\

\textsc{Proto-CLIP}-$F$ & 63.92& 31.32& \textbf{88.55}& 70.35& 78.94& 92.54& 69.59& 62.35& 93.79& 78.29& 74.81& \textbf{33.26} \\

\textsc{Proto-CLIP}-$F$-${Q^{T}}$ & \textbf{64.03}& \textbf{35.82}& 87.46& \textbf{71.50}& \textbf{81.89}& \textbf{92.62}& \textbf{70.02}& \textbf{64.01}& \textbf{94.28}& 78.61& \textbf{75.34}& 32.70 \\\hline

\multicolumn{13}{c}{\textbf{16 shots}} \\ \hline
Linear-Probe CLIP~\cite{radford2021learning} & 55.87 & 36.39 & 76.42 & 70.08 & 82.76 & 90.63 & 67.15 & 63.97 & 94.95 & 70.17 & 73.72 &  -  \\
CLIP-Adapter~\cite{gao2021clip} & 63.59 & 32.10 & 87.84 & 74.01 & 84.43 & 92.49 & 69.55 & 65.96 & 93.90 & 78.25 & 76.76 &  -  \\
CoOp~\cite{zhou2022learning} & 62.95 & 31.26 & 87.01 & 73.36 & 83.53 & 91.83 & 69.26 & 63.58 & 94.51 & 74.67 & 75.71 &  -  \\
Tip~\cite{tip_adapter_eccv22} & 62.02 & 29.76 & 88.14 & 66.77 & 70.54 & 90.18 & 66.85 & 60.93 & 89.89 & 77.83 & 70.58 & 28.87 \\


Tip-F~\cite{tip_adapter_eccv22} & 65.51 & 35.55 & \textbf{89.70} & 75.74 & 84.54 & 92.86 & 71.47 & 66.55 & 94.80 & \textbf{79.43} & 78.03 &  34.04  \\




\textsc{Proto-CLIP} &   62.77& 29.67& 88.61& 68.11& 72.95& 91.08& 68.09& 61.64& 92.94& 78.11& 73.35& 29.96 \\
\textsc{Proto-CLIP}-$F$ &  65.75& 37.56& 89.62 & 75.25& 83.53& 93.43& \textbf{71.94}& \textbf{68.56}& 95.78& 79.09& 77.50& \textbf{35.22} \\

\textsc{Proto-CLIP}-$F$-${Q^{T}}$ & \textbf{65.91}& \textbf{40.65}& 89.34& \textbf{76.76}& \textbf{86.59}& \textbf{93.59}& 72.19 & 68.50& \textbf{96.35}& 79.34 & \textbf{78.11}& 34.70 \\\hline

\hline
\end{tabular}
}
\caption{Few-shot classification results of various CLIP based few shot learning methods on different datasets across various shots using the CLIP ResNet50 backbone.}
\vspace{-2mm}


\label{tab:proto-CLIP_results_new}
\end{table*}

\textbf{Loss functions.} We have introduced three different loss functions in Sec.~\ref{sec: method}: $\mathcal{L}_1, \mathcal{L}_2, \mathcal{L}_3$. We analyze the effects of these loss functions in Table~\ref{tab:loss-ablation}. We can see that i) the $\mathcal{L}_1$ loss function is essential since it drives the classification of the query images; ii) Overall, both $\mathcal{L}_2$ and $\mathcal{L}_3$ loss functions for prototype alignment contribute to the performance, which verifies our motivation of aligning image and text prototypes for few-shot classification.

\begin{table*}[h]
\centering
\resizebox{\textwidth}{!}{%
\begin{tabular}{cccccccccccccc}
\hline
\textbf{Adapter} & \textbf{Train-Text-Memory} & \multicolumn{1}{l}{\textbf{ImageNet}}  & \multicolumn{1}{l}{\textbf{FGVC}}      & \multicolumn{1}{l}{\textbf{Pets}}      & \multicolumn{1}{l}{\textbf{Cars}}      & \multicolumn{1}{l}{\textbf{EuroSAT}}   & \multicolumn{1}{l}{\textbf{Caltech101}} & \multicolumn{1}{l}{\textbf{SUN397}}    & \multicolumn{1}{l}{\textbf{DTD}}       & \multicolumn{1}{l}{\textbf{Flowers}}   & \multicolumn{1}{l}{\textbf{Food101}}   & \multicolumn{1}{l}{\textbf{UCF101}}    & \multicolumn{1}{l}{\textbf{FewSOL}}    \\\hline
MLP               & \textcolor{red}{\ding{55}}                          & 61.06          & 35.31          & 85.61          & 72.19          & 83.47          & 92.58           & 68.54          & 63.89          & 95.01          & 74.05          & 76.16          & 28.65          \\
MLP               & \textcolor{ForestGreen}{\ding{51}}                       & 61.06          & \textbf{37.56} & 85.72          & 73.61          & \textbf{83.53} & 92.13           & 69.71          & 63.89          & \textbf{96.06} & 74.05          & 76.16          & 32.87          \\
2xConv           & \textcolor{red}{\ding{55}}                         & \textsc{\textbf{65.75}} & 34.38          & \textbf{89.62} & \textbf{75.25}          & 81.85          & 93.40           & \textbf{71.94} & 67.85          & 94.76          & \textbf{79.09} & \textbf{77.50}          & 27.13          \\
2xConv           & \textcolor{ForestGreen}{\ding{51}}                        & 58.60          & 35.82          & 89.21          & 74.34         & 81.78          & 93.02           & 69.79          & 67.32 & 95.82          & 78.06          & 76.37          & 27.13 \\
3xConv           & \textcolor{red}{\ding{55}}                          & 65.37          & 34.41           & 88.74          & \textbf{\underline{75.25}} & 82.21          & \textbf{93.43}  & 71.63          & 67.67          & 94.40          & 79.11          & \textbf{\underline{77.50}} & 29.78          \\
3xConv           & \textcolor{ForestGreen}{\ding{51}}                        & 59.63          & 36.15          &    87.93       & 72.68          & 81.57          & 92.74           & 68.64          & \textbf{68.56}          & 95.78          & 78.61          & 77.03          & \textbf{35.22} \\\hline
\end{tabular}
}
\caption{Results of ablation study of various query adapter types and textual memory bank training using the CLIP ResNet50 backbone with $K=16$ on \textsc{Proto-CLIP}-$F$. In case of a tie, the underlined setup was selected randomly.}
\vspace{-2mm}
\label{table:abl-study-adapter-k-16}
\end{table*}

\begin{table*}[h]
\resizebox{\textwidth}{!}{%
\begin{tabular}{ccccccccccccccc}
\hline
\textbf{Loss} & \textbf{ImageNet} & \textbf{FGVC} & \textbf{Pets} & \textbf{Cars} & \textbf{EuroSAT} & \textbf{Caltech101} & \textbf{SUN397} &  \textbf{DTD} & \textbf{Flowers} & \textbf{Food101} & \textbf{UCF101} & \textsc{\textbf{FewSOL}}\\\hline

$\mathcal{L}_1$  & 62.67 &  20.34 &  73.21 &  73.77 &  78.98 &  92.25 &  68.34 &  66.49 &  \textbf{96.14} &  77.39 &  76.66 &  34.57 \\\hline

$\mathcal{L}_2$  & 62.29 &  4.71 &  0.00 &  0.00 &  38.95 &  0.28 &  66.93 &  67.38 &  10.31 &  77.71 &  57.41 &  32.70 \\\hline

$\mathcal{L}_3$  & 62.27 &  4.14 &  0.00 &  0.00 &  38.09 &  0.24 &  64.86 &  67.38 &  10.27 &  77.69 &  57.55 &  20.22 \\\hline

$\mathcal{L}_1 + \mathcal{L}_2$  & 65.39 &  36.24 &  88.58 &  75.39 &  82.78 &  \textbf{93.71} &  71.65 &  68.09 &  96.06 &  78.69 &  77.29 &  33.48 \\\hline

$\mathcal{L}_2 + \mathcal{L}_3$  & 62.33 &  3.87 &  0.00 &  0.00 &  36.86 &  0.24 &  64.84 &  68.32 &  8.20 &  77.35 &  57.52 &  19.61 \\\hline

$\mathcal{L}_1 + \mathcal{L}_3$  & 65.43 &  36.84 &  88.58 &  \textbf{75.51} &  82.84 &  93.35 &  71.44 &  68.32 &  \textbf{96.14} &  78.80 &  \textbf{77.53} &  33.43 \\\hline

$\mathcal{L}_1 + \mathcal{L}_2 + \mathcal{L}_3$  & \textbf{65.75} &  \textbf{37.56} &  \textbf{89.62} &  75.25 &  \textbf{83.53} &  93.43 &  \textbf{71.94} &  \textbf{68.56} &  96.06 &  \textbf{79.09} &  77.50 &  \textbf{35.22} \\\hline

\end{tabular}}
\caption{Ablation study of various Loss functions using the CLIP ResNet50 backbone and $K=16$. The best performing model architectures for each dataset from Table~\ref{table:abl-study-adapter-k-16} are used here.}
\vspace{-7mm}
\label{tab:loss-ablation}
\end{table*}

\begin{figure*}[t]
    \centering
    \includegraphics[width=\textwidth]{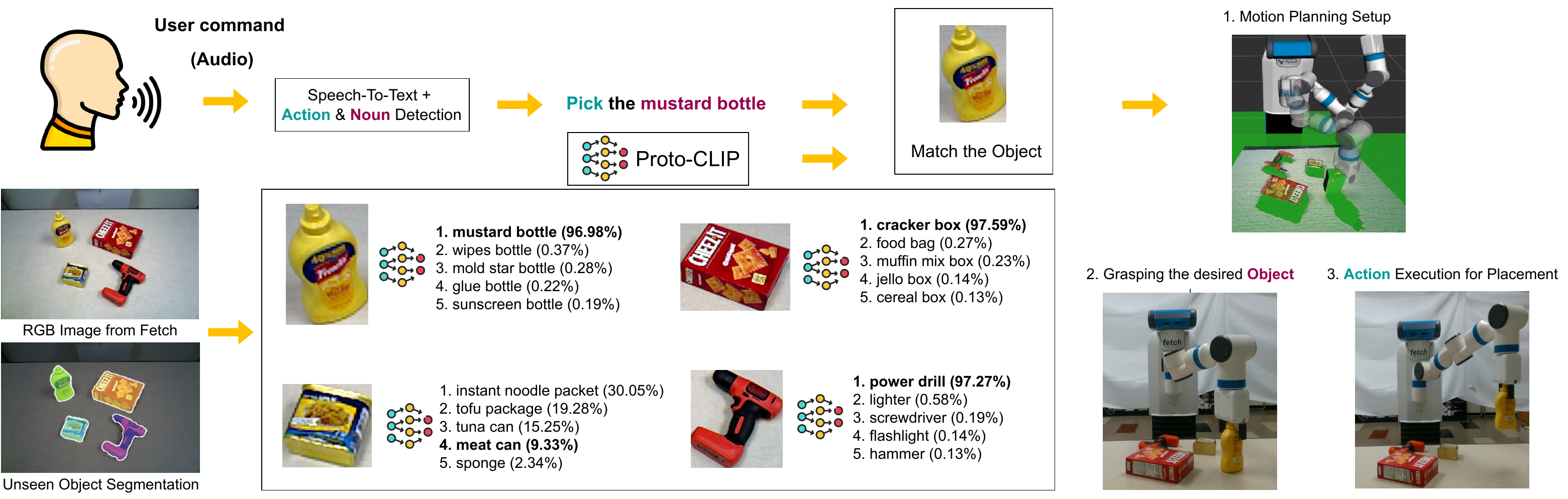}
    \caption{Results for the real world setup with top-5 predictions from the \textsc{Proto-CLIP}-$F$ (ViT-L/14) model trained on \textsc{FewSOL}-198~\cite{p2023fewsol}. The Speech-To-Text is performed via Whisper~\cite{radford2022_whisper}.}
    \label{fig:real-world-exp-setup}
    \vspace{-4mm}
\end{figure*}

\textbf{Backbones.} Table~\ref{tbl: ablation-backbone} shows the results of using different backbone networks on the FewSOL dataset~\cite{p2023fewsol}. In general, better backbones can learn more powerful feature representations and consequently improve the classification accuracy. CLIP vision transformer backbones achieve better performance than CLIP ResNet backbones.

\begin{table}[h]
\resizebox{\linewidth}{!}{%
\begin{tabular}{c|c|c|ccccc}
\hline
\multirow{2}{*}{\textbf{Model}} & \multirow{2}{*}{\textbf{Adapter}} & \multirow{2}{*}{\textbf{TTM}} & \multicolumn{5}{c}{\textbf{Backbone}}                           \\\cline{4-8}
                         &     &                               & \textbf{RN50} & \textbf{RN101} & \textbf{ViT-B/16} & \textbf{ViT-B/32} & \textbf{ViT-L/14} \\\hline
Zero-Shot-CLIP~\cite{radford2021learning}    & -            &  -                               &   25.91   &  32.96     &   40.70        &    41.87      &    54.57    \\\hline
Tip~\cite{tip_adapter_eccv22}                  & -     &  -                               &   29.74   &   37.43    &   47.00        &  41.48        &    56.78      \\
Tip-F~\cite{tip_adapter_eccv22}                   & -    & -                                &   32.52   &   41.43    &     50.17      &   45.48       &    60.17      \\\hline
\textsc{Proto-CLIP}-$F$ & MLP                       & \textcolor{red}{\ding{55}}                                &  33.48    &   39.04    &     47.96      &   41.91       &    58.65      \\
\textsc{Proto-CLIP}-$F$ &MLP                       & \textcolor{ForestGreen}{\ding{51}}   & 34.83     &  40.74     &    47.43    &     42.13     &   58.91             \\
\textsc{Proto-CLIP}-$F$ &2xConv                   & \textcolor{red}{\ding{55}}                                &   35.04   &    41.04         &    50.83      &  46.52        &    \textbf{63.74}      \\
\textsc{Proto-CLIP}-$F$ &2xConv                   & \textcolor{ForestGreen}{\ding{51}}                                &   35.04   &  42.52     &    49.26    &          43.43    &        61.61  \\
\textsc{Proto-CLIP}-$F$ &3xConv                   & \textcolor{red}{\ding{55}}                                &   34.13         &   42.83     &   \textbf{51.91}       &       \textbf{46.87}   &     62.35     \\
\textsc{Proto-CLIP}-$F$ &3xConv                   & \textcolor{ForestGreen}{\ding{51}}                                &  \textbf{35.22}      &    \textbf{44.09}   &     50.39   &           46.57       &      60.39   \\\hline
\end{tabular}
}
\caption{Backbone ablation study. Dataset=\textsc{FewSOL-52}~\cite{p2023fewsol}. $K=16$. TTM=`Train-Text-Memory'.}\label{tbl: ablation-backbone}

\vspace{-2mm}
\end{table}

\begin{table}[h]
\centering
\resizebox{\linewidth}{!}{%
\begin{tabular}{l|l|ccccccc}
\hline


\textbf{Dataset} & \textbf{Method}        & \textbf{1} & \textbf{2} & \textbf{4} & \textbf{8} & \textbf{16} & \textbf{32} & \textbf{64} \\\hline

\multirow{5}{*}{ImageNet~\cite{deng2009imagenet}} & Tip~\cite{tip_adapter_eccv22}  & \textbf{60.70} & \textbf{60.96} & 60.98 & 61.45 & 62.01 & 62.51 & 62.88 \\
&\textsc{Proto-CLIP}             &      60.31     &   60.64        &   \textbf{61.30}        &   \textbf{62.12}         &     \textbf{62.77}        &    \textbf{62.98}    &     \textbf{63.23}     \\\cline{2-9}
& Tip-F~\cite{tip_adapter_eccv22} & \textbf{61.13} & \textbf{61.69} & \textbf{62.52} & 64.00 & 65.51 & 66.58 & \textbf{67.96}  \\
 & \textsc{Proto-CLIP}-$F$ &      60.32      &     60.64       &    61.30        & 63.92           & 65.75            & 66.47       & 65.36        \\
 & \textsc{Proto-CLIP}-$F$-$Q^T$ &      59.12      &     60.48       &    61.80        & \textbf{64.03}           & \textbf{65.91}            & \textbf{66.71}       &     66.90     \\\hline

\multirow{5}{*}{\textsc{FewSOL-52}~\cite{p2023fewsol}} & Tip~\cite{tip_adapter_eccv22}  & \textbf{27.30} & 26.22 & 28.70 & 29.22 & 28.87 & \textcolor{red}{\ding{55}} & \textcolor{red}{\ding{55}}  \\
& \textsc{Proto-CLIP}             &      27.09      &    \textbf{28.35}        &   \textbf{29.13}         &   \textbf{29.83}         &     \textbf{29.96}        & \textcolor{red}{\ding{55}}       & \textcolor{red}{\ding{55}}       \\\cline{2-9}
& Tip-F~\cite{tip_adapter_eccv22} & \textbf{27.91} & \textbf{27.43} & 29.13 & 32.43 & 34.04 & \textcolor{red}{\ding{55}} & \textcolor{red}{\ding{55}} \\
& \textsc{Proto-CLIP}-$F$ &   22.22         &    26.17        &  27.09          &    \textbf{33.26}        &     \textbf{35.22}        & \textcolor{red}{\ding{55}}       & \textcolor{red}{\ding{55}}        \\
 & \textsc{Proto-CLIP}-$F$-$Q^T$ &   21.65   & 25.91 & \textbf{30.30} & 32.70 & 34.70 & \textcolor{red}{\ding{55}} &   \textcolor{red}{\ding{55}}     \\\hline
\end{tabular}
}
\caption{Shots ablation results. Backbone=`CLIP ResNet50'.}\label{tab:shots-ablation}
\vspace{-7mm}
\end{table}



\textbf{Shots.} Table~\ref{tab:shots-ablation} displays the results of using different numbers of shots on ImageNet~\cite{deng2009imagenet} and \textsc{FewSOL}~\cite{p2023fewsol}. With more shots for training, the classification accuracy is improved accordingly. The choice of K=16 for our experiments aligns with the prevalent practice in the field of vision-language few-shot learning. This specific value has been widely adopted, as evidenced in various scholarly works such as ~\cite{gao2021clip, zhou2022learning, tip_adapter_eccv22}
Moreover, given our specific emphasis on the few-shot context, it appeared prudent to exercise caution when surpassing a particular threshold, specifically 16 in our case. As a result, we embarked on an ablation study involving the ImageNet~\cite{deng2009imagenet} dataset. This particular dataset holds the largest number of classes (1000) and thus provided a suitable platform for investigating shots values beyond 16, such as 32 and 64. Despite our intention to explore 128 shots, our experimental hardware's memory limitations prohibited us from pursuing this avenue. Additionally, \textsc{FewSOL} is valuable for few-shot object learning, especially in robotics. We capped shots at 16 for \textsc{FewSOL} as average number of samples per class in \textsc{FewSOL} hovers around 15. Consequently, we conjectured that going beyond might yield diminishing learning returns. These insights are detailed in Table~\ref{tab:shots-ablation}. 

\vspace{-1mm}
\subsection{Comparison with Other Methods}
\label{sec: exp-few-shot-obj-cls}

Table~\ref{tab:proto-CLIP_results_new} shows the performance of \textsc{Proto-CLIP} compared to the state-of-the-art methods using CLIP for few-shot learning in the literature: Linear-Probe CLIP~\cite{radford2021learning}, CoOp~\cite{zhou2022learning}, CLIP-Adapter~\cite{gao2021clip} and Tip-Adapter~\cite{zhou2022learning}. We follow these methods and use CLIP's ResNet50 backbone for this comparison. The fine-tuned variant of Tip-Adapter ``Tip-F'' is the most competitive method compared to ours. The performance of \textsc{Proto-CLIP} on very few shots, i.e., 1 shot and 2 shots is inferior compared to Tip-F. When the number of shots increases to 4, 8 and 16, the fine-tuned variants of \textsc{Proto-CLIP} outperform Tip-F. The enhanced performance of our proposed \textsc{Proto-CLIP} method can be attributed to its reliance on robust image and textual prototypes, which subsequently leads to improved classification accuracy. Therefore, our model benefits from more than 4 shots, while it is not as good as Tip-F when using 1 shot and 2 shots. \textsc{Proto-CLIP}-$F$-$Q^T$ performs better than \textsc{Proto-CLIP}-$F$ on most datasets by using the data augmentation of query images during training\footnote{\label{footnote:suppPDF}For more details, please see the supplementary material on the project page.}.

\subsection{Real World Experiments}\label{sec: real-world-experiments}

As an application, we have built a robotic system to verify the effectiveness of \textsc{Proto-CLIP} for object recognition in the real world. Fig.~\ref{fig:real-world-exp-setup} illustrates our pipeline for the system. It takes human instruction in the form of voice commands as input such as ``pick something'' or ``grasp something''. The system first applies Automatic Speech Recognition (ASR) to convert voice input to text using OpenAI Whisper~\cite{radford2022_whisper}. Then the system grounds the noun in the human instruction into a target object observed from an input image. This is achieved by joint object segmentation and classification. We utilize unseen object instance segmentation~\cite{lu2022mean} to segment objects in cluttered scenes and then classify each segmented object with \textsc{Proto-CLIP}. By matching the noun with the class labels, the system can ground the target in the image. Once the target object is recognized, we use Contact-GraspNet~\cite{sundermeyer2021contactgraspnet} for grasp planning and MoveIt motion planning toolbox~\cite{chitta2012moveit} to pick and place the target\footref{footnote:suppPDF}.

\vspace{-1mm}
\section{CONCLUSIONS} \label{sec:conclusion}
\vspace{-1mm}
We have introduced a novel method for few-shot learning based on the CLIP~\cite{radford2021learning} vision-language model. Our method learns image prototypes and text prototypes from few-shot training examples and aligns the corresponding image-text prototypes for classification. The model is equipped with learnable image memory and text memory for support images and a learnable adapter for query images. Compared to previous CLIP-based few-shot learning methods, our method is flexible in configuring these learnable components, resulting in powerful learned models. Good feature representation is the key in few-shot learning. Future work includes how to further improve feature representation learning compared to CLIP models. One idea is to adapt more powerful vision-language models such as GPT variants. The \textsc{FewSOL}~\cite{p2023fewsol} dataset also provides multiview and depth information about objects. Exploring this 3D information in few-shot object recognition is also a promising direction.







\section*{ACKNOWLEDGMENT}

This work was supported in part by the DARPA Perceptually-enabled Task Guidance (PTG) Program under contract number HR00112220005.

{
\bibliographystyle{IEEEtran}
\bibliography{root}  
}

\end{document}

%% file: title.tex
\textsc{Proto-CLIP}: Vision-Language Prototypical Network for Few-Shot Learning